\documentclass[runningheads]{llncs}
\usepackage[T1]{fontenc}
\usepackage{graphicx}
\usepackage{booktabs}
\usepackage{amsmath,amssymb,amsfonts}

\usepackage{cleveref}
\crefname{section}{Sect.}{Sect.}
\Crefname{section}{Section}{Sections}
\crefname{figure}{Fig.}{Fig.} 
\crefname{table}{Table}{Tables}
\crefname{equation}{Eq.}{Eqs.}

\usepackage[acronym]{glossaries}
\usepackage{glossaries-extra}
\usepackage[super]{nth}
\setabbreviationstyle[acronym]{long-short}
\newacronym{rl}{RL}{reinforcement learning}
\newacronym{bo}{BO}{Bayesian optimization}
\newacronym{moo}{MOO}{multi-objective optimization}
\newacronym{morl}{MORL}{multi-objective reinforcement learning}
\newacronym{mobo}{MOBO}{multi-objective Bayesian optimization}
\newacronym{gp}{GP}{Gaussian process}
\newacronym{qnehvi}{qEHVI}{Expected Hypervolume Improvement}
\newacronym{ppo}{PPO}{Proximal Policy Optimization}
\newacronym{dof}{DoF}{Degree-of-Freedom}
\newacronym{mdp}{MDP}{Markov decision process}

\usepackage{csquotes}

\usepackage{xcolor}
\definecolor{improvement}{RGB}{34,139,34} 
\definecolor{decline}{RGB}{220,20,60} 

\newcommand{\improve}{\textcolor{improvement}{\hfill $\uparrow$}}
\newcommand{\decline}{\textcolor{decline}{\hfill $\downarrow$}}

\usepackage{tikz}
\usetikzlibrary{shapes.geometric, arrows.meta, positioning, fit, backgrounds}

\begin{document}

\title{Sample-Efficient Pareto Front Modeling for Energy-Aware Reinforcement Learning Using Bayesian Optimization}

\titlerunning{Bayesian Pareto Modeling for RL}

\author{Georg Schäfer\inst{1,2} \and Jakob Rehrl\inst{1} \and Stefan Huber\inst{1} \and Simon Hirlaender\inst{2}}

\authorrunning{G. Schäfer et al.}

\institute{Josef Ressel Centre for Intelligent and Secure Industrial Automation,\\
Salzburg University of Applied Sciences, Salzburg, Austria \and
Department of Artificial Intelligence and Human Interfaces,\\
Paris Lodron University of Salzburg, Salzburg, Austria\\
\email{georg.schaefer@fh-salzburg.ac.at}}

\maketitle

\begin{abstract}
Industrial automation increasingly demands control strategies that balance operational performance with strict energy efficiency requirements. A common approach to solving this multi-objective problem, particularly within the framework of \gls{rl}, is to formulate a single, scalar reward function that linearly combines the competing objectives. However, the manual weighting of these different objectives is heavily reliant on domain intuition, incredibly time-consuming, prone to human bias, and frequently fails to uncover optimal trade-off solutions. This work addresses the critical challenge of automating the weight selection process to systematically and efficiently discover the Pareto front of optimal trade-off policies. We formulate the weight selection process as a \gls{mobo} problem and evaluate its sample efficiency against a standard uniform grid search baseline. Using a physical Quanser Aero 2 testbed configured for 1-\gls{dof} pitch control, our results demonstrate that the \gls{mobo} approach, utilizing the Expected Hypervolume Improvement (qEHVI) acquisition function, consistently outperforms uniform grid sampling. \gls{mobo} achieves superior hypervolume and maximum spread, successfully identifying high-quality, diverse trade-off policies with a reduced evaluation budget, thereby enabling highly efficient energy-aware control in complex mechatronic systems.
\end{abstract}

\section{Introduction}
\label{sec:intro}
The rapid modernization of industrial automation has brought forward a critical design paradigm: systems must no longer be optimized solely for operational performance, but must concurrently adhere to strict energy efficiency and sustainability constraints \cite{biel2020energy}. Traditional control methodologies, while robust, often struggle to adapt to the highly nonlinear dynamics, complex environmental interactions, and shifting objectives present in modern mechatronic systems. Consequently, \gls{rl} has emerged as a powerful, data-driven framework capable of learning optimal control policies through continuous interaction with an environment \cite{kober2013reinforcement}. Indeed, recent empirical evaluations have demonstrated that policy gradient methods can offer highly competitive adaptability and rapid response characteristics when benchmarked against traditional Model Predictive Control (MPC) on non-linear mechatronic systems \cite{SRHH24}.

However, integrating energy efficiency into the \gls{rl} framework fundamentally transforms the control task into a multi-objective optimization problem.
The agent must balance at least two inherently conflicting goals: maximizing primary operational performance, such as precise setpoint regulation, dynamic trajectory tracking, or process throughput, and minimizing the physical energy expended during actuation.
Traditional approaches frequently treat energy consumption as a secondary heuristic via fixed, manually tuned reward penalties.
While our approach utilizes a similar linear scalarization to combine competing objectives, we address the critical limitation of manual tuning by formulating a framework that automates the discovery of the optimal trade-off surface \cite{roijers2013survey}.

In industrial practice, this selection is predominantly achieved through manual tuning. A control engineer intuitively selects a weight, trains a complete \gls{rl} agent, evaluates the physical performance, and iteratively adjusts the weight based on the observed outcome. This methodology is incredibly time-consuming, highly susceptible to human bias, and frequently fails to uncover the true Pareto front, i.e., the set of optimal solutions where no single objective can be improved without degrading another \cite{deb2014multi}. Because training an \gls{rl} agent on physical hardware or high-fidelity simulators is computationally expensive and subjects machinery to physical wear, there is a pressing need for sample-efficient automation of this process.

To address these challenges, this paper proposes an automated framework for modeling the optimal trade-off surface in energy-aware continuous control. Our main contributions are as follows:
\begin{itemize}
    \item We formulate the multi-objective reward weighting process as a black-box optimization problem and apply \gls{mobo} to automate the discovery of the Pareto front.
    \item We conduct a rigorous empirical comparison between the proposed \gls{mobo} approach and a standard uniform grid search baseline, utilizing a high-fidelity environment of a Quanser Aero 2 mechatronic testbed.
    \item We demonstrate that \gls{mobo} achieves superior Pareto front approximations across multiple quantitative metrics (hypervolume, spacing, maximum spread) while drastically reducing the required computational budget and environment interactions.
\end{itemize}

The remainder of this paper is structured as follows. \Cref{sec:background} provides the necessary theoretical background and discusses related work in the fields of multi-objective RL and Bayesian optimization. \Cref{sec:system_formulation} details the system architecture of the mechatronic testbed and formally defines the reinforcement learning problem. \Cref{sec:methodology} outlines our implementation and evaluation methodology. Finally, \Cref{sec:results} discusses the empirical results, followed by concluding remarks and directions for future research in \Cref{sec:conclusion}.

\section{Background and Related Work}
\label{sec:background}

\subsection{Multi-Objective Reinforcement Learning}
Standard \gls{rl} operates on the premise of maximizing a single scalar reward signal, traditionally modeled as a \gls{mdp} \cite{sutton2018reinforcement}. However, industrial applications inherently involve multiple conflicting feedback signals. As Roijers et al.~\cite{roijers2013survey} establish, converting a multi-objective problem into a single-objective one via fixed scalarization is often undesirable when the end-user's specific performance-versus-energy preferences are unknown a priori. In such scenarios, the objective shifts from finding a single optimal policy to identifying the set of Pareto-optimal policies. Recent literature emphasizes that explicitly multi-objective methods are required to solve complex planning problems efficiently \cite{hayes2022practical}. Let the performance of a policy $\pi$ be evaluated by a vector of objective functions $\mathbf{J}(\pi) = [J_1(\pi), J_2(\pi), \dots, J_k(\pi)]^T$. A policy $\pi^*$ is considered Pareto optimal if there does not exist another policy $\pi$ such that $J_i(\pi) \leq J_i(\pi^*)$ for all objectives $i$, and $J_j(\pi) < J_j(\pi^*)$ for at least one objective $j$ \cite{miettinen1998nonlinear}. The set of all such non-dominated policy evaluations forms the Pareto front.

\subsection{Energy-Aware Control in Industrial Automation}
In modern industrial robotics, balancing system performance, such as precise trajectory tracking or manipulation, with energy efficiency is critical. Traditional approaches frequently treat energy consumption as a secondary heuristic or a simple reward penalty. However, sustainable operation demands that actuation energy is explicitly regulated. Recent frameworks address this by treating the task as a Constrained \gls{mdp} or a multi-objective problem, seeking to minimize long-term energy budgets while preserving strict operational reliability \cite{liu2023energyaware}. Our approach aligns with this paradigm by directly modeling the trade-off between tracking error and power consumption, focusing specifically on automating the discovery of the optimal trade-off surface.

\subsection{Bayesian Optimization for Pareto Front Modeling}
To map the Pareto front without exhaustively training \gls{rl} policies for every possible weight combination, sample-efficient black-box optimization is necessary. \gls{bo} \cite{shahriari2015taking} has become a standard approach for optimizing expensive functions, utilizing a \gls{gp} \cite{rasmussen2006gaussian} surrogate model to approximate the objective landscape and quantify uncertainty. In the multi-objective context (\gls{mobo}), Gaussian Processes are combined with specialized acquisition functions, such as Expected Hypervolume Improvement (qEHVI) \cite{daulton2020differentiable}, to intelligently select query points that maximally expand the non-dominated solution set. Recent advancements focus on efficiently evaluating Pareto-frontier entropy \cite{suzuki2020multi} and selecting diverse batches of parameters to ensure broad coverage \cite{ahmadianshalchi2024pareto}. Our work directly applies these \gls{mobo} principles to automate the hyperparameter weighting of an \gls{rl} reward function.

\section{System Architecture and Problem Formulation}
\label{sec:system_formulation}

\subsection{The Quanser Aero 2 Testbed}
\begin{figure}[ht]
    \centering
    \includegraphics[width=\linewidth]{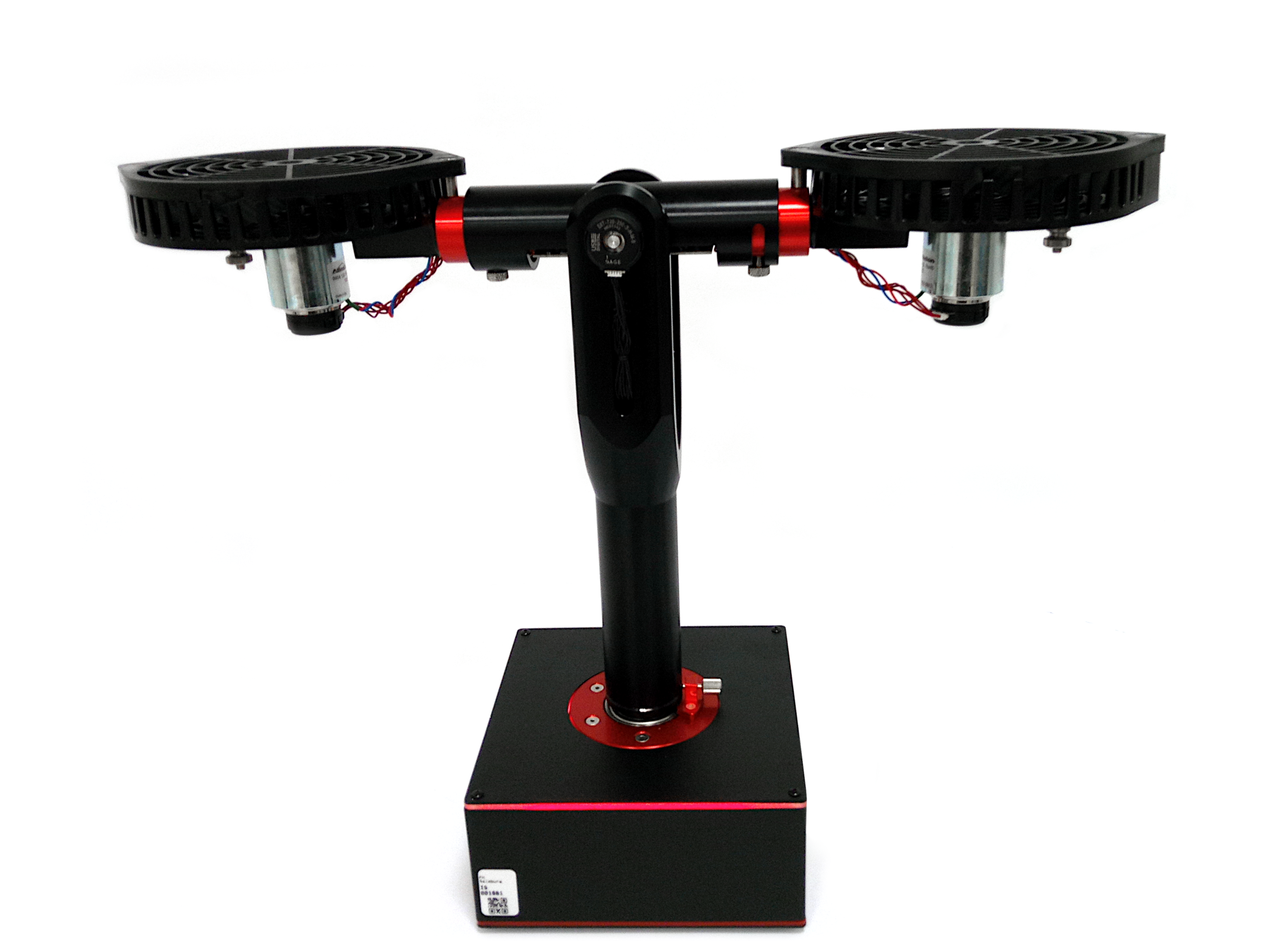}
    \caption{The Quanser Aero 2 mechatronic testbed. For this study, the system is configured in a 1-DoF mode where the yaw axis is locked and the pitch axis is controlled via the dual DC motors.}
    \label{fig:aero_testbed}
\end{figure}

To rigorously validate our approach, we require a mechatronic system featuring non-linear dynamics and measurable physical energy states. We utilize the Quanser Aero 2 system (\cref{fig:aero_testbed}). For the scope of this evaluation, the testbed is configured in a 1-\gls{dof} setup: the yaw axis is mechanically locked, while the pitch axis remains unlocked and free to rotate.

The system is actuated by two DC motors equipped with propellers. Let $u_0$ and $u_1$ represent the voltage signals applied to the respective motors. In the 1-\gls{dof} pitch configuration, the control input $u$ is symmetric and defined as $u = u_0 = -u_1$. The primary output $y = \Theta$ represents the pitch angle of the main beam. The system dynamics are inherently non-linear due to the aerodynamic drag, friction, and thrust characteristics of the propellers, providing an established, highly representative benchmark for evaluating advanced learning-based control algorithms \cite{SRHH24}.

\subsection{State and Action Space Formulation}
The task assigned to the \gls{ppo} agent \cite{schulman2017proximal} is to orient the beam to a desired, time-varying target angle while minimizing the electrical power consumed by the DC motors. The state space $\mathcal{S}$ is fully observable and represented as a continuous vector containing the current pitch angle, the angular velocity, and the current target reference angle:
\begin{equation}
    s_t = \left[ \Theta_t, \dot{\Theta}_t, \Theta^{ref}_t \right]^T
\end{equation}
The action space $\mathcal{A}$ comprises the continuous voltage signal applied to the motors, bounded by the physical limits of the hardware:
\begin{equation}
    a_t = [ u_t ] \in [-24\text{V}, 24\text{V}]
\end{equation}

\subsection{Reward Function Design}
The behavior of the agent is entirely dictated by the parameter $\alpha$ in the scalarized reward function. At each time step $t$, the reward $R_t$ is computed as:
\begin{equation}
    R_t = -(1-\alpha) \cdot |\Delta_t| - \alpha \cdot P_t
    \label{eq:scalar_reward}
\end{equation}
To ensure stable neural network updates and meaningful scalarization, the terms $\Delta_t$ (absolute tracking error) and $P_t$ (power consumption) are normalized to a $[0, 1]$ scale.
As demonstrated in our prior investigations regarding the formulation of RL environments, strict normalization of observation spaces and actions is a crucial prerequisite for stable policy convergence and the successful transfer of simulated policies to the physical system \cite{SKRHH25}. The power consumption $P_t$ is calculated dynamically based on the applied voltage and the internal resistance characteristics of the Aero 2 motors. By manipulating $\alpha \in [0, 1]$ in our previous work \cite{SSRHH25}, we observed that the problem spans the entire spectrum from maximum tracking performance to maximum energy conservation, validating its use as the core tunable hyperparameter.

\section{Methodology}
\label{sec:methodology}

\subsection{Software and Implementation Details}
The \gls{rl} environment was built as a high-fidelity digital twin of the physical Quanser Aero 2 system to ensure that all trained policies are highly deployable. This environment utilizes our previously established Python-Simulink co-simulation framework \cite{SSRHH24}, which enables the seamless integration of numerically robust Simulink models with flexible Python-based machine learning environments. The \gls{ppo} \cite{schulman2017proximal} agent was implemented utilizing the Stable Baselines3 framework \cite{stable-baselines3}, backed by PyTorch \cite{paszke2019pytorch} for automatic differentiation and neural network optimization. Both the actor and critic networks feature continuous multi-layer perceptron (MLP) architectures. 

\subsection{MOBO-RL Integration Workflow}
To automate the discovery of the Pareto front, we integrate the \gls{rl} training process within an outer-loop optimization framework managed by the BoTorch/Ax library~\cite{olson2025ax}. The optimization is initialized by defining the continuous parameter bounds $\alpha \in [0.0, 1.0]$ and the two minimization objectives: tracking error ($f_1$) and power consumption ($f_2$). 

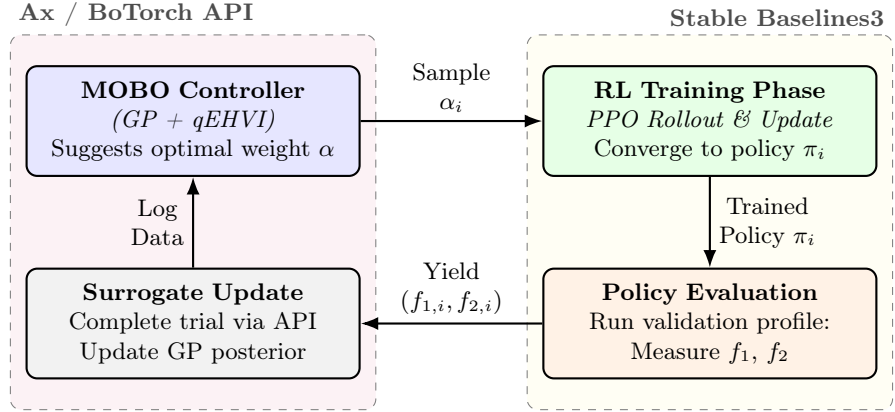
\begin{figure}[ht]
\centering
\begin{tikzpicture}[
    node distance=1.5cm and 2.4cm,
    box/.style={rectangle, draw=black, thick, text width=4.0cm, align=center, rounded corners, font=\small, inner sep=0.2cm},
    arrow/.style={-Latex, thick},
    label/.style={font=\footnotesize, align=center}
]

\node[box, fill=blue!10] (mobo) {\textbf{MOBO Controller}\\ \textit{(GP + qEHVI)}\\ Suggests optimal weight $\alpha$};

\node[box, right=2.4cm of mobo, fill=green!10] (rl) {\textbf{RL Training Phase}\\ \textit{PPO Rollout \& Update}\\ Converge to policy $\pi_i$};

\node[box, below=1.2cm of rl, fill=orange!10] (eval) {\textbf{Policy Evaluation}\\ Run validation profile:\\ Measure $f_1$, $f_2$};

\node[box, fill=gray!10] (update) at (mobo |- eval) {\textbf{Surrogate Update}\\ Complete trial via API\\ Update GP posterior};

\draw[arrow] (mobo) -- node[above, label] {Sample\\$\alpha_i$} (rl);
\draw[arrow] (rl) -- node[right, label] {Trained\\Policy $\pi_i$} (eval);
\draw[arrow] (eval) -- node[above, label] {Yield\\$(f_{1,i}, f_{2,i})$} (update);
\draw[arrow] (update) -- node[left, label] {Log\\Data} (mobo);

\begin{scope}[on background layer]
\node[draw=black!50, dashed, inner xsep=0.2cm, inner ysep=0.4cm, fit=(rl)(eval), fill=yellow!5, rounded corners] (innerloop) {};
\node[above left, font=\footnotesize\bfseries, color=black!70] at (innerloop.north east) {Stable Baselines3};

\node[draw=black!50, dashed, inner xsep=0.2cm, inner ysep=0.4cm, fit=(mobo)(update), fill=purple!5, rounded corners] (outerloop) {};
\node[above right, font=\footnotesize\bfseries, color=black!70] at (outerloop.north west) {Ax / BoTorch API};
\end{scope}

\end{tikzpicture}
\caption{The proposed MOBO-RL integration workflow. The Bayesian optimizer iteratively samples the reward weighting parameter $\alpha$. For each sampled weight, an \gls{rl} agent is trained to convergence and its physical performance is evaluated. The resulting empirical observations update the surrogate model to intelligently guide future sampling.}
\label{fig:mobo_workflow}
\end{figure}

The structure of this framework is illustrated in \cref{fig:mobo_workflow}. The surrogate model approximates the objective functions using a \gls{gp}. For each sequential trial, the optimizer queries this surrogate model to suggest the next most promising $\alpha$ weight utilizing the \gls{qnehvi} acquisition function. A fresh \gls{ppo} agent is instantiated and trained to convergence using this specific scalarized reward formulation. Following training, the resultant policy $\pi$ is deployed in a deterministic evaluation environment to record the true tracking error and electrical power draw. Finally, these empirical results are passed back to the optimizer via the API, dynamically updating the posterior distribution of the surrogate model and informing the subsequent iteration.

\subsection{Exploration Strategies}
To approximate the Pareto front, we enforce a strict, heavily constrained maximum budget of 11 full \gls{rl} training runs (evaluations). This budget simulates the real-world constraints of hardware training times. We compare two distinct strategies for exploring the parameter space of $\alpha$:
\begin{itemize}
    \item \textbf{Uniform Grid Search (Baseline):} We manually select 11 equally spaced values for $\alpha \in [0, 1]$. While providing simple, uniform coverage of the parameter space, this method is fundamentally inefficient as it cannot adapt to the underlying non-linear mapping between $\alpha$ and the bi-objective performance.
    \item \textbf{Multi-Objective BO (MOBO):} To prevent premature convergence and provide a baseline statistical variance, the \gls{mobo} optimization is initialized with 5 quasi-random Sobol points. This is followed by 6 sequentially guided \gls{bo} selections. The total remains exactly 11 trials to ensure a mathematically fair comparison against the baseline.
\end{itemize}

\subsection{Quantitative Evaluation Metrics}
We quantitatively assess the quality of the generated Pareto fronts using three standard multi-objective metrics. Let $\mathcal{P}$ be the approximated Pareto front.
\begin{enumerate}
    \item \textbf{Hypervolume (HV):} Evaluates the total dominated n-dimensional volume in the objective space between the Pareto solutions and a predefined, worst-case reference point $\mathbf{r}$. 
    \begin{equation}
        \text{HV}(\mathcal{P}) = \Lambda \left( \bigcup_{\mathbf{x} \in \mathcal{P}} [f_1(\mathbf{x}), r_1] \times [f_2(\mathbf{x}), r_2] \right)
    \end{equation}
    where $\Lambda$ denotes the Lebesgue measure. A larger volume indicates a front that is closer to the true Pareto optimal set and possesses a better spread.
    \item \textbf{Spacing ($S$):} Measures the uniformity of the solution distribution along the front. It is defined as the variance in the Euclidean distance between adjacent solutions:
    \begin{equation}
        S = \sqrt{\frac{1}{n-1} \sum_{i=1}^{n} (\bar{d} - d_i)^2}
    \end{equation}
    where $d_i$ is the distance to the nearest neighbor for the $i$-th solution, and $\bar{d}$ is the mean of all $d_i$. Lower values indicate a more evenly distributed set of choices.
    \item \textbf{Maximum Spread ($MS$):} Assesses the maximal range coverage by calculating the Euclidean distance between the extreme boundary points of the discovered front in the objective space. Let $K$ be the number of objectives. The maximum spread is defined as the Euclidean norm of the objective ranges:
    \begin{equation}
        MS = \sqrt{\sum_{k=1}^{K} \left( \max_{\mathbf{x} \in \mathcal{P}} f_k(\mathbf{x}) - \min_{\mathbf{x} \in \mathcal{P}} f_k(\mathbf{x}) \right)^2}
    \end{equation}
    A higher $MS$ value indicates that the algorithm successfully explored the absolute outer limits of the performance trade-off, providing a broader spectrum of operational modes.
\end{enumerate}

\section{Results and Discussion}
\label{sec:results}

\subsection{Quantitative Results}
\Cref{tab:result} summarizes the final performance of both exploration strategies after the strict budget of 11 evaluations was exhausted. The automated \gls{mobo} approach consistently outperforms the uniform grid sampling across all three defined multi-objective metrics. 

\begin{table}[ht]
\centering
\setlength{\tabcolsep}{10pt} 
\caption{Comparison of Pareto front quality metrics after 11 trials.}
\label{tab:result}
\begin{tabular}{lccc}
\toprule
Metric & Grid Search & \gls{mobo} & Difference \\
\midrule
Hypervolume & 762.78 \decline & \textbf{769.48} \improve & +6.71 \\
Spacing & 12.26 \decline & \textbf{11.22} \improve & -1.04 \\
Maximum Spread & 51.31 \decline & \textbf{51.33} \improve & +0.02 \\
\bottomrule
\end{tabular}
\end{table}

\begin{figure}[ht]
    \centering
    \includegraphics[width=\linewidth]{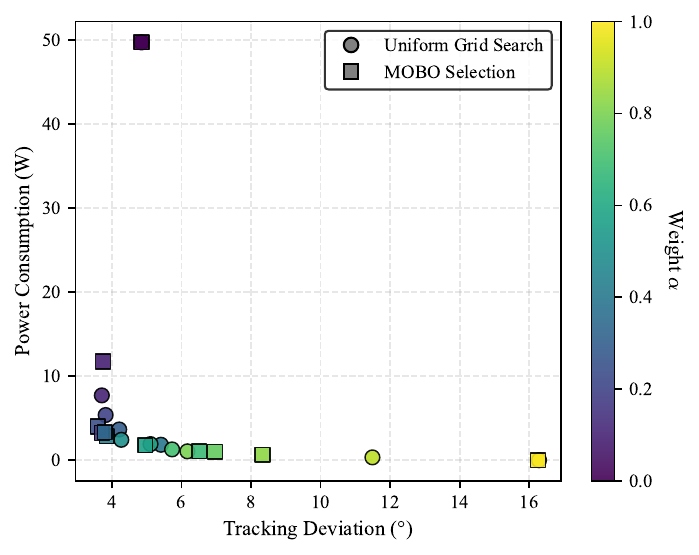}
    \caption{Comparison of all sampled configurations between the uniform grid search and MOBO selection, highlighting the discovered Pareto front.}    \label{fig:pareto_front}
\end{figure}

The higher Hypervolume demonstrates that \gls{mobo} discovered policies that objectively dominate those found by the uniform grid, as visualized in \cref{fig:pareto_front}. Crucially, the non-linear relationship between the weight $\alpha$ and the actual physical system performance causes a uniform grid search in the \textit{parameter} space to map to a highly non-uniform cluster of points in the \textit{objective} space. As observed in our experiments, the Grid Search wasted a significant portion of its 11 evaluations on adjacent $\alpha$ values that resulted in similar physical system behavior, largely failing to map the critical \enquote{knee} of the Pareto front where the most viable trade-offs reside.

\subsection{Sample Efficiency and Progression}

\begin{figure}[ht]
    \centering
    \includegraphics[width=\linewidth]{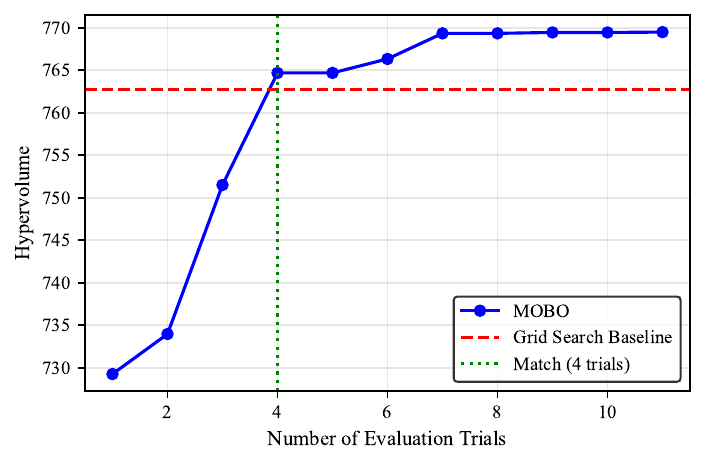}
    \caption{Progression of hypervolume across the evaluation trials compared to the grid search baseline.}
    \label{fig:hypervolume_progression}
\end{figure}

The most significant and practical advantage of the \gls{mobo} approach is its remarkable sample efficiency, which directly translates to saved computational time and reduced hardware degradation during physical training. Analyzing the progression of the hypervolume (\cref{fig:hypervolume_progression}), \gls{mobo} rapidly identifies high-quality solutions. Because the qEHVI acquisition function explicitly targets areas of the objective space that maximize expected improvement, it intelligently avoids redundant regions of the parameter space.

Remarkably, the \gls{mobo} approach achieves the final, maximum hypervolume of the 11-trial Grid Search baseline in just 4 evaluation trials. In a real-world industrial context, where training a single policy to convergence might take hours or days of continuous operation, matching baseline performance in 4 trials represents a reduction in computational and temporal cost of approximately 64\%. 

\begin{figure}[ht]
    \centering
    \includegraphics[width=\linewidth]{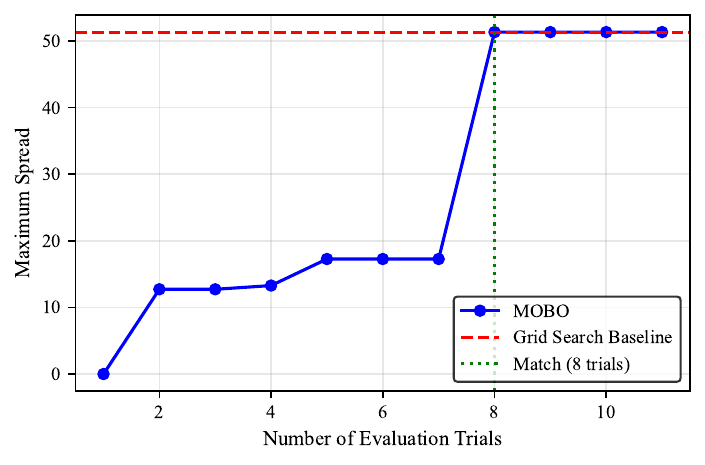}
    \caption{Progression of maximum spread across the evaluation trials compared to the grid search baseline.}
    \label{fig:spread_progression}
\end{figure}

Similarly, the maximum spread progression (\cref{fig:spread_progression}) highlights the capacity of \gls{bo} for active boundary exploration. The algorithm matched the grid search's maximum spread by trial 8. This proves that \gls{mobo} does not merely exploit a single high-performing region, but actively pushes the boundaries of the known objective space. It ensures that fundamentally diverse trade-off policies, spanning from ultra-precise, high-energy trajectory tracking to loose, highly conservative energy-saving modes, are systematically discovered and made available to the control engineer without requiring tedious manual intervention.

\section{Conclusion and Future Work}
\label{sec:conclusion}
This work rigorously validates Multi-Objective Bayesian Optimization as a highly effective, sample-efficient, and automated method for multi-objective weight selection in industrial \gls{rl} continuous control tasks. By successfully balancing electrical energy consumption and tracking performance on the Quanser Aero 2 testbed, we demonstrated that \gls{mobo} fundamentally eliminates the biases and inefficiencies of manual heuristic weighting. The approach consistently creates a superior approximation of the Pareto front while significantly reducing the required evaluation budget, marking a critical step toward accessible, sustainable intelligent automation.

Future work will focus on extending this methodology to the full, highly coupled 2-\gls{dof} (pitch and yaw) configuration of the Quanser Aero 2 to handle more complex trajectory tracking tasks. Furthermore, because exploring the boundaries of the weight parameter space, specifically where tracking performance is heavily prioritized over energy conservation ($\alpha \to 0$), can lead to aggressive, potentially hardware-damaging \enquote{bang-bang} policies, we intend to integrate strict Safe RL constraints into the optimization framework. This ensures that boundary exploration remains physically safe during early training phases. Finally, we will investigate the transferability of these findings, utilizing the identified Pareto optimal weights from the Aero 2 to jump-start and accelerate \gls{rl} training processes on similar, yet mathematically distinct, industrial mechatronic systems.

\subsubsection*{Acknowledgments}
Financial support for this study was provided by the Christian Doppler Research Association (CDG) through the Josef Ressel Centre for Intelligent and Secure Industrial Automation, the corresponding WISS Co-project of Land Salzburg, and by the European Interreg project BA0100172 AI4GREEN.

\bibliographystyle{splncs04}
\bibliography{references,jrcisia-published}

\end{document}